\documentclass{article}

\usepackage{microtype}
\usepackage{booktabs} 

\usepackage[pagebackref=true,breaklinks=true,colorlinks,bookmarks=false]{hyperref}
\usepackage{url}            
\usepackage{booktabs}       
\usepackage{amsfonts}       
\usepackage{nicefrac}       
\usepackage{xcolor}         
\usepackage{multirow}
\usepackage{dsfont}
\usepackage{soul}
\usepackage{adjustbox}
\usepackage{xspace}
\usepackage{bm}
\usepackage{amsmath}
\usepackage{setspace}                       
\usepackage{scrextend}
\usepackage{mathtools}
\usepackage{graphicx}
\usepackage{subcaption}
\usepackage{enumitem}
\usepackage{colortbl}
\usepackage{wrapfig}
\usepackage{kotex}
\usepackage[pagebackref=true,breaklinks=true,colorlinks,bookmarks=false]{hyperref}

\makeatletter
\DeclareRobustCommand\onedot{\futurelet\@let@token\@onedot}
\def\@onedot{\ifx\@let@token.\else.\null\fi\xspace}
\def\eg{\emph{e.g}\onedot} 
\def\ie{\emph{i.e}\onedot}

\newcommand{\trans}{{\scriptscriptstyle\mathsf{T}}}

\usepackage[accepted]{icml2021}

\definecolor{pinegreen}{rgb}{0.0, 0.47, 0.44}
\definecolor{cornellred}{rgb}{0.7, 0.11, 0.11}
\definecolor{cadmiumgreen}{rgb}{0.0, 0.42, 0.24}

\hypersetup{
  linkcolor = cornellred,
  citecolor  = cadmiumgreen,
  colorlinks = true,
}

\newcommand{\llname}{Logic-guided Generation\xspace} 
\newcommand{\lname}{logic-guided generation\xspace} 
\newcommand{\sname}{LoGe\xspace} 

\usepackage{etoolbox}
\usepackage[textsize=scriptsize,textwidth=2.2cm]{todonotes}

\newtoggle{showtodos}
\toggletrue{showtodos} 

\iftoggle{showtodos}{
\newcommand{\swmo}[1]{\todo[color=magenta]{SW: #1}}
\newcommand{\sungs}[1]{\todo[color=cadmiumgreen]{SS: #1}}
\newcommand{\sh}[1]{\todo[color=blue]{SH: #1}}
}{
\newcommand{\swmo}[1]{}
\newcommand{\sungs}[1]{}
\newcommand{\sh}[1]{}
}

\begin{document}
\twocolumn[
\icmltitle{Abstract Reasoning via \llname}
\icmltitlerunning{Abstract Reasoning via \llname}

\icmlsetsymbol{equal}{*}
\begin{icmlauthorlist}
\icmlauthor{Sihyun Yu}{ka}
\icmlauthor{Sangwoo Mo}{ka}
\icmlauthor{Sungsoo Ahn}{mu}
\icmlauthor{Jinwoo Shin}{ka}
\end{icmlauthorlist}

\icmlaffiliation{ka}{Korea Advanced Institute of Science and Technology (KAIST)}
\icmlaffiliation{mu}{Mohamed bin Zayed University of Artificial Intelligence (MBZUAI)}

\icmlcorrespondingauthor{Sihyun Yu}{sihyun.yu@kaist.ac.kr}
\vskip 0.3in
]
\printAffiliationsAndNotice{} 

\begin{abstract}
Abstract reasoning, \ie, inferring complicated patterns from given observations, is a central building block of artificial general intelligence. While humans find the answer by either eliminating wrong candidates or first constructing the answer, prior deep neural network (DNN)-based methods focus on the former discriminative approach. This paper aims to design a framework for the latter approach and bridge the gap between artificial and human intelligence. To this end, we propose \lname (\sname), a novel generative DNN framework that reduces abstract reasoning as an optimization problem in propositional logic. \sname is composed of three steps: extract propositional variables from images, reason the answer variables with a logic layer, and reconstruct the answer image from the variables. We demonstrate that \sname outperforms the black box DNN frameworks for generative abstract reasoning under the RAVEN benchmark, \ie, reconstructing answers based on capturing correct rules of various attributes from observations.
\end{abstract}

\section{Introduction}
\label{sec:intro}
\begin{figure*}[t]
\centering
    \begin{subfigure}{0.28\linewidth}
        \centering
        \includegraphics[width=\textwidth]{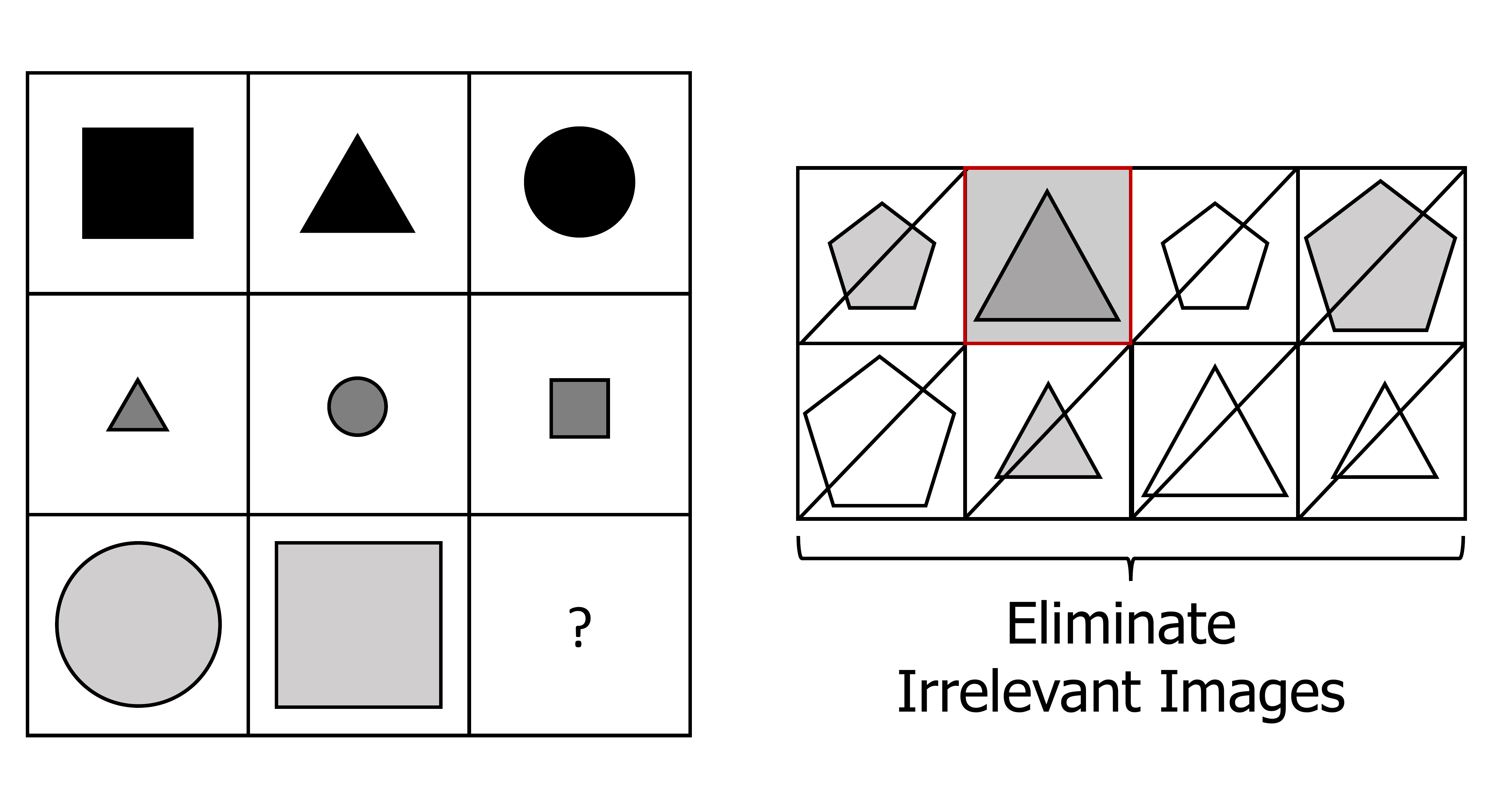}
        \caption{Response elimination.}
        \label{fig:1_elimination}
    \end{subfigure}%
    ~
    \begin{subfigure}{0.42\linewidth}
        \centering
        \includegraphics[width=\textwidth]{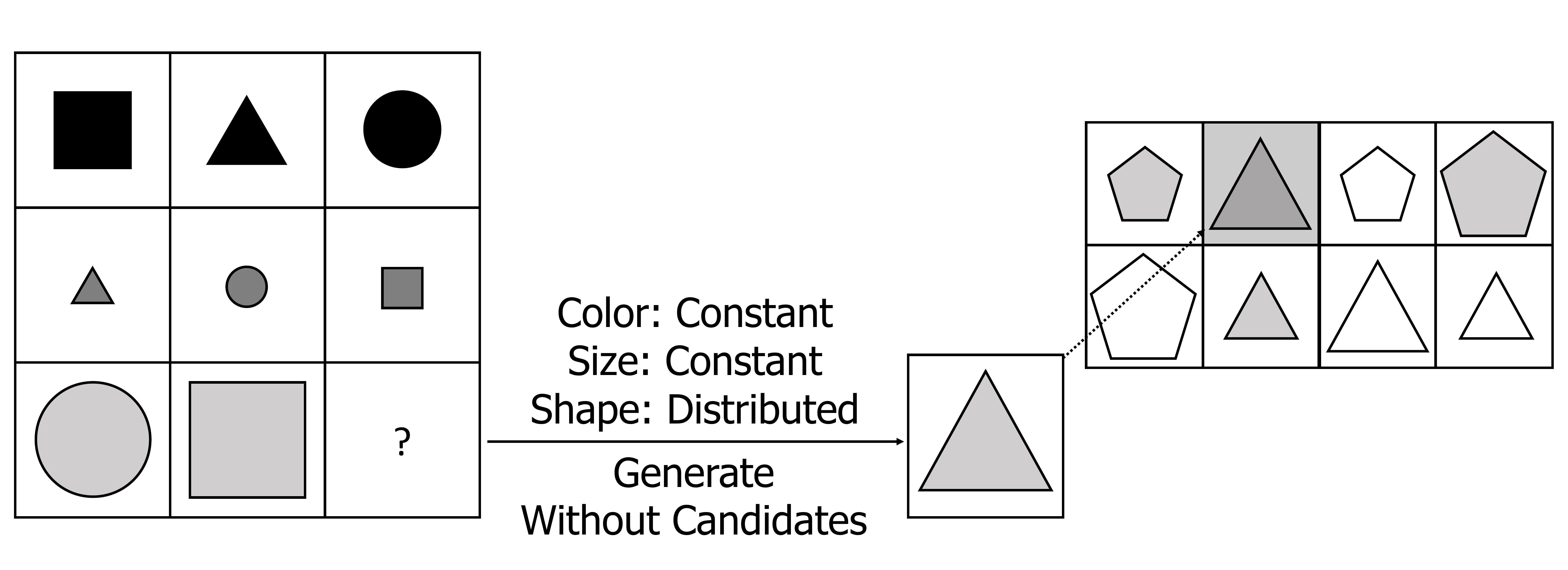}
        \caption{Constructive matching.}
        \label{fig:1_matching}
    \end{subfigure}
    ~
    \begin{subfigure}{0.72\linewidth}
        \centering
        \includegraphics[width=\textwidth]{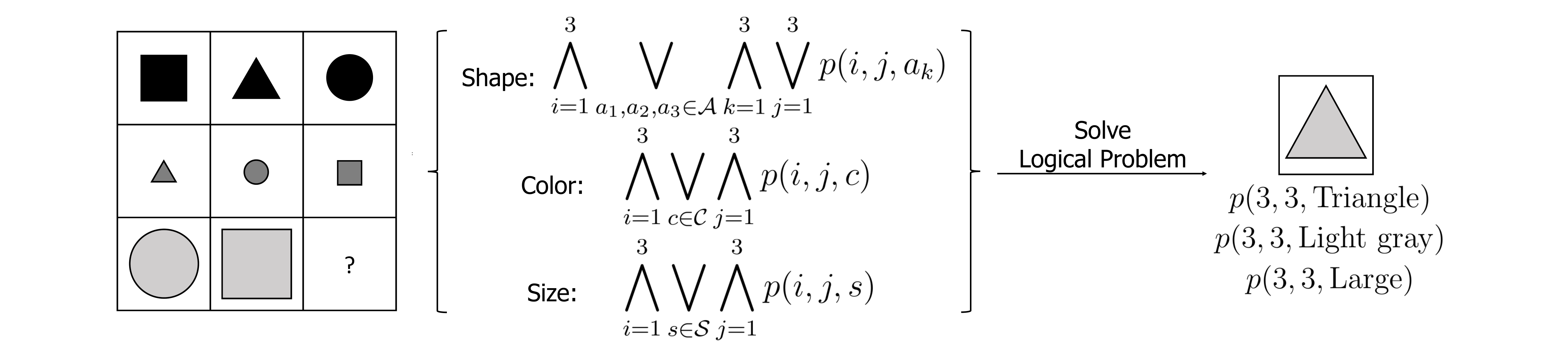}
        \caption{Connection to propositional logic.}
        \label{fig:1_propositional}
    \end{subfigure}
    \caption{Illustration of strategies to solve Raven's Progressive Matrices (RPM) problem. (a) In response elimination, one finds the answer by eliminating the irrelevant candidates which are not aligned from context images. (b) In constructive matching, one first imagines the answer candidate from inferred rules then selects the answer. (c) Interpretation of the problem combined with the propositional logic.}
\label{fig:1_strategy}
\vspace{-0.15in}
\end{figure*}

Imitating the human ability to infer complicated patterns from observations has been a long-standing goal of artificial intelligence. In order to build such models capable of this reasoning ability, recent works \citep{zheng2019abstract, zhang2019learning, hahne2019attention, wang2020abstract, hu2020hierarchical, wu2020scattering} have focused on training a deep neural network (DNN) which solves abstract reasoning problems that resemble an IQ test (Figure~\ref{fig:1_strategy}). In these problems, one should infer common rules of contexts without any additional information other than context images and select a correct answer from a candidate set. Accordingly, those DNN-based approaches attempt to derive the framework in situations where both the supervision on rules of the problem and the explicit symbol labels of each image are not provided. A couple of studies \citep{santoro2018measuring, zhang2019raven} demonstrated how the widely used neural network architectures such as ResNet \citep{he2016deep} or LSTM \citep{hochreiter1997long} are unfit for learning reasoning capability, as any priors to resemble the human reasoning procedure are not employed in these architectures. Remarkably, recent works have shown that the performance can be significantly improved with the careful neural architecture design motivated by the human reasoning process and even outperforms humans \citep{zhang2019learning, zheng2019abstract, wu2020scattering}.
 
Most DNN-based methods on abstract reasoning mostly have resembled how humans perform reasoning via \emph{response elimination} strategy (Figure~\ref{fig:1_elimination}), \ie, to exclude candidate answers based on matching with the given context images. Intriguingly, cognitive science \citep{bethell1984adaptive, carpenter1990one} reveals that humans use two types of abstract reasoning strategies, not only limited to response elimination strategy. To be specific, humans also have the ability to perform \emph{constructive matching} (Figure~\ref{fig:1_matching}); they can first imagine the answer from context images without any candidates and then match the candidate answers to select the most similar one. Especially, several works \citep{mitchum2010solve, becker2016preventing} have emphasized that the latter strategy better reflects the general intelligence of humans. However, investigation on the ability of neural networks to achieve constructive matching is yet under-explored; even such a direction is promising.

\textbf{Contribution.} We introduce a new end-to-end generative framework, coined \emph{\lname (\sname)}, to learn a constructive matching strategy on abstract reasoning like humans. Our main idea is to reduce these reasoning problems into optimization problems in propositional logic. Leveraging such prior knowledge, \sname learns to embed each image to discrete variables and generate the answer image via incorporating a differentiable propositional logical reasoning layer. We note that both objectives are achieved without any supervision on the exact propositional variables of each image and underlying rules in the problem. Specifically, we propose a three-step framework to achieve these objectives: abstraction, reasoning, and reconstruction. 

We verify how our \sname effectively solves the proposed task on the RAVEN \citep{zhang2019raven} dataset. To be specific, we show that our framework generates high-quality images, which is correct, based on capturing the underlying abstract rules and attributes. This result is remarkable, as the widely used neural network architectures perform poorly for this conditional generation task. We also verify how \sname performs comparably to neural networks that rely on response elimination strategy to perform abstract reasoning, even though our task is arguably harder and has not accessed to the wrong candidates while training.

\section{\llname}
\label{sec:method}

In this section, we demonstrate \textit{\lname} (\sname), a framework to imitate a human's constructive matching strategy on abstract reasoning. 

\subsection{Overview of \lname}
Our problem setup is largely inspired by \citet{bethell1984adaptive}, who evaluated the constructive matching ability of humans to measure their generative reasoning ability. In this perspective, we express reasoning as a task of inferring the rule $r$ from a given problem $P$, where the problem $P$ is a pair of a context $\bm{x}$ and answer $y$ satisfying the rule $y=r(\bm{x}) $. We especially focus on the generative strategy for solving this task; given a query context $\bm{x}$, we evaluate the ability of machines to infer a rule as $\hat{r}$ from contexts $\bm{x}$ to generate an answer image $\hat{y}=\hat{r}({\bm{x}})$ that matches the ground-truth image $y$.

For teaching models the ability of generative strategy in abstract reasoning, we train them on a dataset $\mathcal{D}$ consisting of $d$ problems, \ie, $\mathcal{D} = \{({\bm{x}}_{i}, y_{i})\}_{i=1}^{d}$. To be specific, we consider a dataset where each context $\bm{x}$ in the dataset $\mathcal{D}$ is a tuple of $M$ images $(x_1, \ldots, x_M)$ and $y$ is an answer image. Images are specified by a collection of abstract features such as shapes, colors and size. For instance, we visualize the case of the generative strategy in Raven’s Progressive Matrices (RPM) structure in Figure \ref{fig:1_matching}: contexts are given as eight images \ie, $M=8$, and the goal is to generate an answer image for the remaining location, denoted by a question mark.

Our main idea is to connect abstract reasoning to the optimization problem in propositional logic to achieve the generative reasoning strategy into the framework. For instance, in RPM problems, one can define propositional variables $p(i, j, o)$ as ``an image placed at $i$th row and $j$th column contains an attribute $o$'' to represent contexts in the problem, as shown in Figure \ref{fig:1_propositional}. Here, attributes are sets of features in each context image, \eg, set of shapes $\mathcal{A}$, color $\mathcal{C}$, and size $\mathcal{S}$. With those variables, underlying rules can be written as propositional logical formulas as in Figure \ref{fig:1_propositional}. In this respect, one may interpret the answer generation procedure as the MAXSAT optimization problem in propositional logic: finding propositional variables representing the answer which satisfy the underlying logical formula in the given RPM problem as much as possible. We provide a more description of the MAXSAT problem in Appendix \ref{appen:maxsat}.

\begin{figure*}[t]
\begin{center}
\includegraphics[scale=0.34]{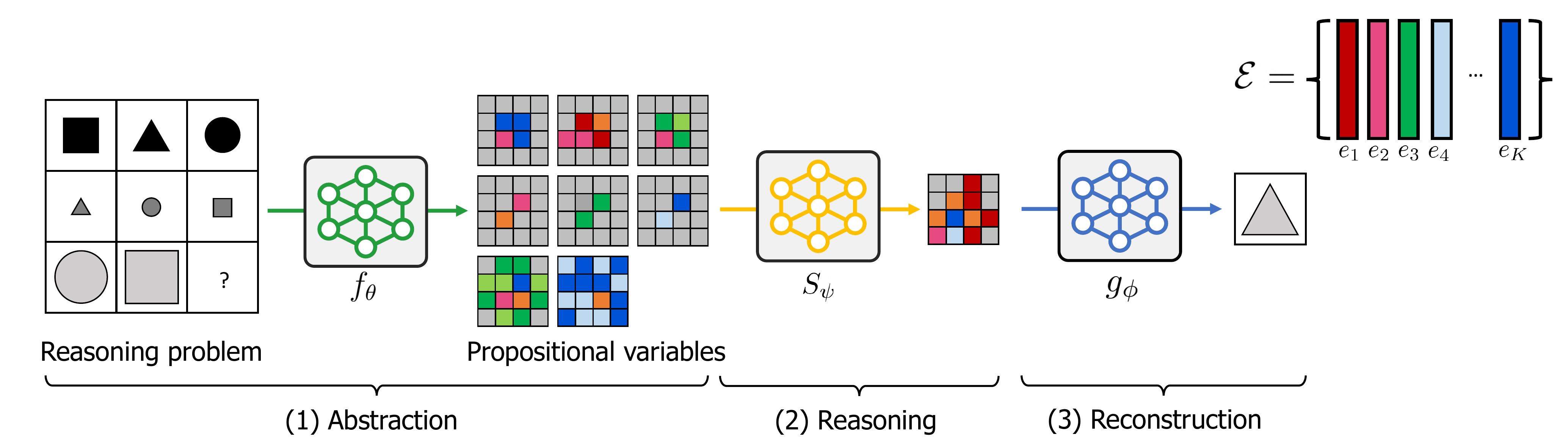}
\end{center}
\caption{Overall illustration of our framework consists of three steps: abstraction, reasoning, and reconstruction. We compute the propositional logical variables of each image and predict variables of the answer combined with the logical reasoning layer.} 
\label{fig:2_method}
\vspace{-0.2in}
\end{figure*}

As those propositional variables are not provided in dataset $\mathcal{D}$, \sname learns a propositional embedding and rules in problems in self-supervised manner. In particular, we derive a three-step framework with an encoder network $f_\theta$, a decoder network $g_{\phi}$, a logical reasoning layer $S_{\psi}$ parametrized by $\theta$, $\phi$, $\psi$, respectively, and a latent codebook $\mathcal{E} = \{e_1,\cdots, e_K\}$, where each element $e_k \in \mathcal{E}$ is a trainable $D$-dimensional real-valued vector:
\vspace{-0.1in}
\begin{itemize}
    \item \textbf{(Abstraction.)} The encoder network $f_{\theta}$ and the codebook $\mathcal{E}$ embeds contexts into propositional variables.
    \item \textbf{(Reasoning.)} The reasoning layer $S_{\psi}$ predicts propositional variables of the answer image.
    \item \textbf{(Reconstruction.)} The decoder network $g_{\phi}$ and the codebook $\mathcal{E}$ generates the answer image from inferred propositional variables. 
\end{itemize}
We provide an illustration of our framework in Figure \ref{fig:2_method}.

\subsection{Detailed description of \sname}
In the rest of this section, we describe each step of our framework in detail. 

\textbf{Abstraction.}
We first compute propositional logical variables of each context image $x_i \in \bm{x}$ from the encoder network $f_\theta$ and the latent codebook $\mathcal{E}$. 
To achieve this, we first pass through each image $x_i$ into the encoder network $f_\theta$ to have a corresponding output $z_i \coloneqq f_\theta(x_i) \in \mathbb{R}^{H\times D}$, in which we denote $z_i=(z_{(i,1)},\ldots, z_{(i,H)})^{\trans}$. 
We then quantize the output $z_i$ with the codebook $\mathcal{E}$, denoted by $q_i\coloneqq(e_{s(i,1)},\ldots, e_{s(i,H)})^{\trans}$, where $s(i,h)\in[K]$ with $[K]\coloneqq\{1,\ldots,K\}$ is defined as follows:
\begin{align*}
s(i,h) &\coloneqq \underset{k \in [K]}{\arg \min}\,\, \lVert z_{(i,h)} -e_k \rVert_2, \quad \forall h \in [H], 
\end{align*}
where $[H]\coloneqq \{ 1,\ldots, H\}$. We finally consider an one-hot encoding of indices of $q_i$, namely $(s(i,1), \ldots, s(i, H))$. Specifically, we map each index $s(i,h)\in[K]$ into $K$-categorical one-hot vector $t_{(i,H)}$ in which the $s(i,h)$-th value is 1 for all $h \in [H]$. Consequently, we have an one-hot embedding $t_i \coloneqq (t_{(i,1)},\ldots,t_{(i,H)})^{\trans} \in \{0,1\}^{H \times K}$ of each image $x_i$. 
This one-hot embedding $t_i$ is regarded and utilized as propositional variables of $x_i$ in further steps.

\textbf{Reasoning.} 
With propositional variables $(t_1, \ldots, t_M)$ of the context $\bm{x}$, we compute propositional variables $\hat{t}\in\{ 0,1\}^{H \times K}$ which corresponds to the predicted answer image. 
To be specific, we evaluate $\hat{t}$ from the reasoning layer $S_\phi$ and propositional variables of contexts $\bm{x}$, \ie, $\hat{t} = S_\psi(t_1, \ldots, t_M)$.
For the reasoning layer $S_{\psi}$, we choose the SATNet layer \cite{wang2019satnet}, which is a differentiable version of the MAXSAT problem solver and learns propositional logical formulas from data as layer weights. 
We provide details of this reasoning layer in Appendix \ref{appen:satnet}.

\textbf{Reconstruction.} 
Finally, we infer the answer image $\hat{y}$ from predicted propositional variables $\hat{t} \in\{ 0,1\}^{H \times K}$. To achieve this, we first compute a latent vector $\hat{q}\in \mathbb{R}^{H\times D}$ of $\hat{y}$ from predicted $\hat{t}$ and the codebook $\mathcal{E}$:
\begin{align*}
\hat{q} \coloneqq \hat{t}E,\quad \text{where}\,\, E = (e_1, \ldots, e_K)^{\trans}\in\mathbb{R}^{K\times D}.
\end{align*}
We then return the output from the decoder $g_{\phi}(\hat{q})$ as the final answer image $\hat{y}$.

\textbf{Training objective.}
To train \sname, we propose three loss functions: $\mathcal{L}_{\mathtt{abst}}$, $\mathcal{L}_{\mathtt{reason}}$, and $\mathcal{L}_{\mathtt{recon}}$ for each $(\bm{x}, y) \in \mathcal{D}$, which is for abstraction, reasoning, and reconstruction step, respectively. To formulate those objectives, we additionally consider $q^{\ast}$ and $t^{\ast}$, which indicates a quantized vector and propositional variables of the answer $y$ from the abstraction step.

We first formulate $\mathcal{L}_{\mathtt{abst}}$ and $\mathcal{L}_{\mathtt{recon}}$, which resemble the objective in vector-quantized variational autoencoder \citep{van2017neural, razavi2019generating}:
\begin{figure*}[t]
\centering
\includegraphics[width=0.52\linewidth]{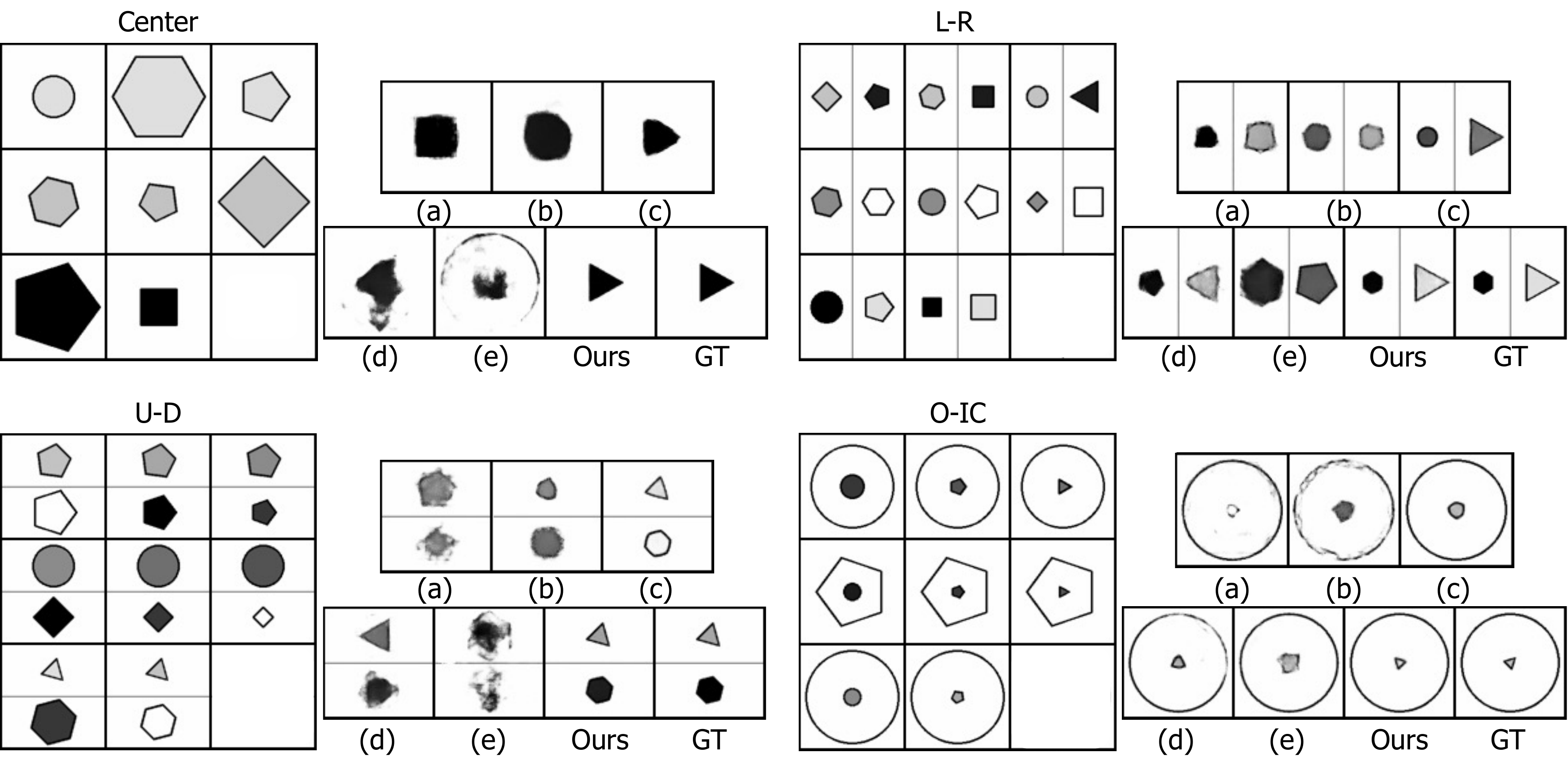}
\caption{Comparison of generated and ground truth image on RAVEN dataset among different architecture variants. (a)-(c): autoencoder with an attention layer and two convolutional networks with different kernel sizes as the reasoning network. We set the kernel size of 3 and 1 for (b) and (c), respectively. (d)-(e): VQ-VAE with an attention layer and a convolutional network with a kernel size as 3. Ours: our \sname framework, which is composed of VQ-VAE and SATNet. GT: ground-truth answer image.} 
\label{fig:3_main}
\vspace{-0.1in}
\end{figure*}
\begin{align*}
    \mathcal{L}_{\mathtt{abst}}
    \coloneqq&\sum_{x_i \in \bm{x}} 
    \Big[
    \lVert \overline{f_{\theta} (x_i)} - q_i 
    \rVert_2^{2} + \beta\lVert f_{\theta} (x_i) - \overline{q_i} \rVert_2^{2} \Big] \\
    &+\lVert \overline{f_{\theta} (y)} - q^{\ast} \rVert_2^{2} 
    + \beta\lVert f_{\theta} (y) - \overline{q^{\ast}} \rVert_2^{2}, \\
    \mathcal{L}_{\mathtt{recon}} \coloneqq& 
   \sum_{x_i \in \bm{x}} \lVert g_\phi (q_i) - x_i||_2^2 + \rVert g_\phi (q^{\ast}) - y \rVert_2^2,
   \end{align*}
where the term with the bar indicates the term with a stop-gradient operator. Moreover, we define $\mathcal{L}_{\mathtt{reason}}$ as follows:
\begin{align*}
        \mathcal{L}_{\mathtt{reason}} \coloneqq \mathcal{L}_{\mathtt{BCE}} (S_\psi(t_1, \ldots, t_M), t^{\ast}).
\end{align*}
Here, $\mathcal{L}_{\mathtt{BCE}}$ denotes the binary cross-entropy loss. 

To sum up, we optimize the loss $\mathcal{L}_{\mathtt{total}}(\theta, \phi, \psi, \mathcal{E})$, which is a sum of above three loss functions:
\begin{align*}
        \mathcal{L}_{\mathtt{total}}(\theta, \phi, \psi, \mathcal{E}) \coloneqq 
        \mathcal{L}_{\mathtt{abst}} +  \mathcal{L}_{\mathtt{recon}} +
        \mathcal{L}_{\mathtt{reason}}.
\end{align*}
Here, the total loss $\mathcal{L}_{\mathtt{total}}$ contains several discrete outputs, \eg, outputs of the reasoning layer in $\mathcal{L}_{\mathtt{reason}}$. We provide a detailed description of how to deal with such non-differentiability on the optimization in Appendix \ref{appen:optimize}.

\vspace{-0.1in}
\section{Experiments}
\label{sec:exp}
We verify the effectiveness of our framework on the RAVEN/i-RAVEN dataset \citep{zhang2019raven, hu2020hierarchical}. Our result demonstrates that the proposed \lname (\sname) framework how well generates the answer image at a given abstract reasoning problem, while other neural architectures fail to achieve this. Moreover, we also show our framework can be employed for discriminative tasks, \ie, choose the answer among candidates, and it shows improved results compared to existing discriminative methods \citep{zhang2019raven, zheng2019abstract, zhang2019learning, hu2020hierarchical, wu2020scattering}.
\begin{table}[t]
\centering
\small
\resizebox{\linewidth}{!}{
\begin{tabular}{cccccc}
    \toprule
    Method & Center & U-D & L-R & O-IC \\
    \midrule
    LSTM \citep{zhang2019raven} & 12.3 & 10.3 & 12.7 & 12.9 \\
    WReN \citep{santoro2018measuring} & 23.3 & 15.2 & 16.5 & 16.8  \\
    LEN \citep{zheng2019abstract} & 42.5 &
    28.1 & 27.6 & 32.9 \\
    CoPINet \citep{zhang2019learning} & 50.4 & 40.8 & 40.0 & 42.7 \\
    SRAN \citep{hu2020hierarchical} & 53.4 & 43.1 & 41.4 & 44.0\\
    \midrule
    \sname (Ours) & \textbf{87.5} & \textbf{64.0} & \textbf{51.7} & \textbf{48.5} \\
    \bottomrule
\end{tabular}
}
\caption{Accuracy of discriminate tasks on i-RAVEN dataset from existing approaches and \sname. For \sname, we select the image with the smallest mean-squared error from the generated image.}
\label{tab:1_quantitative}
\vspace{-0.30in}
\end{table}

\vspace{-0.05in}
\subsection{Experimental setup} 
\textbf{Datasets.}
To verify the effectiveness of \sname, we choose RAVEN \citep{zhang2019raven} and i-RAVEN \citep{hu2020hierarchical}. For more details of these datasets, see Appendix \ref{appen:raven}.

\textbf{Baselines.}
We note that \sname utilizes VQ-VAE \citep{van2017neural} and SATNet \citep{wang2019satnet} to leverage a propositional logic. To verify the effectiveness of this logical prior, we qualitatively compare the generated answers with ones from other black-box neural network frameworks, \ie, different encoders and reasoning networks other than VQ-VAE and SATNet, respectively. For quantitative results, we compare the performance with prior methods on discriminative abstract reasoning. We provide detailed descriptions of baselines for comparison in Appendix \ref{appen:baselines}.

\vspace{-0.05in}
\subsection{Main results}
\textbf{Qualitative result.}
Figure \ref{fig:3_main} summarizes results of the generated answers from different configurations of RAVEN dataset. \sname successfully generates the answer image at the various configuration of abstract reasoning problems, while other black-box architecture design choices fail. It indicates how propositional logic prior is beneficial to achieve the objective. We provide more illustrations in Appendix \ref{appen:gen}.

\textbf{Quantitative result.}
Table \ref{tab:1_quantitative} shows the comparison of our framework and prior approaches on discriminative tasks.\footnote{For \citet{hu2020hierarchical}, we resized the image to 80$\times$80 for a fair comparison with other baselines.} We select the answer by choosing the candidate which has the smallest mean-squared error from the generated answer image to employ our framework into discriminative tasks. Intriguingly, \sname shows better performance to existing works; even our method has not been accessed to other candidates other than the answer while training.

\vspace{-0.1in}
\section{Conclusion}
\vspace{-0.05in}
\label{sec:conclusion}
We introduce a new deep neural network generative framework that resembles a human’s constructive matching strategy on abstract reasoning. Specifically, we derive a three-step procedure based on connecting the optimization problem in propositional logic and abstract reasoning. Experimental results demonstrate the effectiveness of our framework among various problem types in abstract reasoning to generate the correct answer based on capturing common patterns with propositional logic prior.

\section{Acknowledgements}
This work was partially supported by Institute of Information \& Communications Technology Planning \& Evaluation (IITP) grant funded by the Korea government (MSIT) 
(No.2019-0-00075, Artificial Intelligence Graduate School Program (KAIST)).
This work was mainly supported by Samsung Research Funding \& Incubation Center of Samsung Electronics under Project Number SRFCIT1902-06.

\bibliography{egbib}

\begin{thebibliography}{24}
\providecommand{\natexlab}[1]{#1}
\providecommand{\url}[1]{\texttt{#1}}
\expandafter\ifx\csname urlstyle\endcsname\relax
  \providecommand{\doi}[1]{doi: #1}\else
  \providecommand{\doi}{doi: \begingroup \urlstyle{rm}\Url}\fi

\bibitem[Barvinok(1995)]{barvinok1995problems}
Barvinok, A.~I.
\newblock Problems of distance geometry and convex properties of quadratic
  maps.
\newblock \emph{Discrete \& Computational Geometry}, 13\penalty0 (2), 1995.

\bibitem[Becker et~al.(2016)Becker, Schmitz, Falk, Feldbr{\"u}gge, Recktenwald,
  Wilhelm, Preckel, and Spinath]{becker2016preventing}
Becker, N., Schmitz, F., Falk, A.~M., Feldbr{\"u}gge, J., Recktenwald, D.~R.,
  Wilhelm, O., Preckel, F., and Spinath, F.~M.
\newblock Preventing response elimination strategies improves the convergent
  validity of figural matrices.
\newblock \emph{Journal of Intelligence}, 4\penalty0 (1):\penalty0 2, 2016.

\bibitem[Bengio et~al.(2013)Bengio, L{\'e}onard, and
  Courville]{bengio2013estimating}
Bengio, Y., L{\'e}onard, N., and Courville, A.
\newblock Estimating or propagating gradients through stochastic neurons for
  conditional computation.
\newblock \emph{arXiv preprint arXiv:1308.3432}, 2013.

\bibitem[Bethell-Fox et~al.(1984)Bethell-Fox, Lohman, and
  Snow]{bethell1984adaptive}
Bethell-Fox, C.~E., Lohman, D.~F., and Snow, R.~E.
\newblock Adaptive reasoning: Componential and eye movement analysis of
  geometric analogy performance.
\newblock \emph{Intelligence}, 8\penalty0 (3):\penalty0 205--238, 1984.

\bibitem[Carpenter et~al.(1990)Carpenter, Just, and Shell]{carpenter1990one}
Carpenter, P.~A., Just, M.~A., and Shell, P.
\newblock What one intelligence test measures: a theoretical account of the
  processing in the raven progressive matrices test.
\newblock \emph{Psychological review}, 97\penalty0 (3):\penalty0 404, 1990.

\bibitem[Goemans \& Williamson(1995)Goemans and
  Williamson]{goemans1995improved}
Goemans, M.~X. and Williamson, D.~P.
\newblock Improved approximation algorithms for maximum cut and satisfiability
  problems using semidefinite programming.
\newblock \emph{Journal of the ACM (JACM)}, 42\penalty0 (6), 1995.

\bibitem[Hahne et~al.(2019)Hahne, L{\"u}ddecke, W{\"o}rg{\"o}tter, and
  Kappel]{hahne2019attention}
Hahne, L., L{\"u}ddecke, T., W{\"o}rg{\"o}tter, F., and Kappel, D.
\newblock Attention on abstract visual reasoning.
\newblock \emph{arXiv preprint arXiv:1911.05990}, 2019.

\bibitem[He et~al.(2016)He, Zhang, Ren, and Sun]{he2016deep}
He, K., Zhang, X., Ren, S., and Sun, J.
\newblock Deep residual learning for image recognition.
\newblock In \emph{Proceedings of the IEEE conference on computer vision and
  pattern recognition}, pp.\  770--778, 2016.

\bibitem[Hochreiter \& Schmidhuber(1997)Hochreiter and
  Schmidhuber]{hochreiter1997long}
Hochreiter, S. and Schmidhuber, J.
\newblock Long short-term memory.
\newblock \emph{Neural computation}, 9\penalty0 (8):\penalty0 1735--1780, 1997.

\bibitem[Hu et~al.(2021)Hu, Ma, Liu, Wei, and Bai]{hu2020hierarchical}
Hu, S., Ma, Y., Liu, X., Wei, Y., and Bai, S.
\newblock Stratified rule-aware network for abstract visual reasoning.
\newblock In \emph{AAAI conference on Artificial Intelligence}, 2021.

\bibitem[Mitchum \& Kelley(2010)Mitchum and Kelley]{mitchum2010solve}
Mitchum, A.~L. and Kelley, C.~M.
\newblock Solve the problem first: Constructive solution strategies can
  influence the accuracy of retrospective confidence judgments.
\newblock \emph{Journal of Experimental Psychology: Learning, Memory, and
  Cognition}, 36\penalty0 (3):\penalty0 699, 2010.

\bibitem[Pataki(1998)]{pataki1998rank}
Pataki, G.
\newblock On the rank of extreme matrices in semidefinite programs and the
  multiplicity of optimal eigenvalues.
\newblock \emph{Mathematics of operations research}, 23\penalty0 (2), 1998.

\bibitem[Razavi et~al.(2019)Razavi, van~den Oord, and
  Vinyals]{razavi2019generating}
Razavi, A., van~den Oord, A., and Vinyals, O.
\newblock Generating diverse high-fidelity images with vq-vae-2.
\newblock In \emph{Advances in Neural Information Processing Systems}, 2019.

\bibitem[Santoro et~al.(2017)Santoro, Raposo, Barrett, Malinowski, Pascanu,
  Battaglia, and Lillicrap]{NIPS2017_e6acf4b0}
Santoro, A., Raposo, D., Barrett, D.~G., Malinowski, M., Pascanu, R.,
  Battaglia, P., and Lillicrap, T.
\newblock A simple neural network module for relational reasoning.
\newblock In \emph{Advances in Neural Information Processing Systems}, 2017.

\bibitem[Santoro et~al.(2018)Santoro, Hill, Barrett, Morcos, and
  Lillicrap]{santoro2018measuring}
Santoro, A., Hill, F., Barrett, D., Morcos, A., and Lillicrap, T.
\newblock Measuring abstract reasoning in neural networks.
\newblock In \emph{International Conference on Machine Learning}, 2018.

\bibitem[Van Den~Oord et~al.(2017)Van Den~Oord, Vinyals, et~al.]{van2017neural}
Van Den~Oord, A., Vinyals, O., et~al.
\newblock Neural discrete representation learning.
\newblock In \emph{Advances in Neural Information Processing Systems}, 2017.

\bibitem[Wang et~al.(2020)Wang, Jamnik, and Lio]{wang2020abstract}
Wang, D., Jamnik, M., and Lio, P.
\newblock Abstract diagrammatic reasoning with multiplex graph networks.
\newblock In \emph{International Conference on Learning Representations}, 2020.

\bibitem[Wang \& Kolter(2019)Wang and Kolter]{wang2018low}
Wang, P.-W. and Kolter, J.~Z.
\newblock Low-rank semidefinite programming for the max2sat problem.
\newblock In \emph{AAAI Conference on Artificial Intelligence}, 2019.

\bibitem[Wang et~al.(2017)Wang, Chang, and Kolter]{wang2017mixing}
Wang, P.-W., Chang, W.-C., and Kolter, J.~Z.
\newblock The mixing method: low-rank coordinate descent for semidefinite
  programming with diagonal constraints.
\newblock \emph{arXiv preprint arXiv:1706.00476}, 2017.

\bibitem[Wang et~al.(2019)Wang, Donti, Wilder, and Kolter]{wang2019satnet}
Wang, P.-W., Donti, P.~L., Wilder, B., and Kolter, Z.
\newblock Satnet: Bridging deep learning and logical reasoning using a
  differentiable satisfiability solver.
\newblock In \emph{International Conference on Machine Learning}, 2019.

\bibitem[Wu et~al.(2020)Wu, Dong, Grosse, and Ba]{wu2020scattering}
Wu, Y., Dong, H., Grosse, R., and Ba, J.
\newblock The scattering compositional learner: Discovering objects,
  attributes, relationships in analogical reasoning.
\newblock \emph{arXiv preprint arXiv:2007.04212}, 2020.

\bibitem[Zhang et~al.(2019{\natexlab{a}})Zhang, Gao, Jia, Zhu, and
  Zhu]{zhang2019raven}
Zhang, C., Gao, F., Jia, B., Zhu, Y., and Zhu, S.-C.
\newblock Raven: A dataset for relational and analogical visual reasoning.
\newblock In \emph{Proceedings of the IEEE Conference on Computer Vision and
  Pattern Recognition}, 2019{\natexlab{a}}.

\bibitem[Zhang et~al.(2019{\natexlab{b}})Zhang, Jia, Gao, Zhu, Lu, and
  Zhu]{zhang2019learning}
Zhang, C., Jia, B., Gao, F., Zhu, Y., Lu, H., and Zhu, S.-C.
\newblock Learning perceptual inference by contrasting.
\newblock In \emph{Neural Information Processing Systems}, 2019{\natexlab{b}}.

\bibitem[Zheng et~al.(2019)Zheng, Zha, and Wei]{zheng2019abstract}
Zheng, K., Zha, Z.-J., and Wei, W.
\newblock Abstract reasoning with distracting features.
\newblock In \emph{Advances in Neural Information Processing Systems}, 2019.

\end{thebibliography}
\bibliographystyle{icml2021}
\clearpage

\onecolumn
\appendix
\section{Detailed description of the MAXSAT problem}
\label{appen:maxsat}
\textbf{CNF formula.}
Conjunctive normal form (CNF) formula is a conjunction of clauses, where each clause is composed of OR operation of propositional variables. To be specific, the following is an example of the CNF formula with 2 propositional variables $p_1$ and $p_2$:
\begin{gather}
    (p_1 \vee p_2) \wedge (\neg p_1 \vee p_2) \wedge (p_1 \vee \neg  p_2) \wedge (\neg p_1 \vee \neg p_2).
    \label{eq:cnf}
\end{gather}
\textbf{MAXSAT problem.}
MAXSAT problem aims to find values of propositional variables to maximize the number of clauses of given propositional logical formula in CNF. One may interpret the MAXSAT problem as a combinatorial optimization problem. Specifically, by letting $P:=[p_1, p_2, \cdots, p_n] \in \{-1, 1\}^n$ as $n$ propositional variables and $S: = [s_1, s_2 \cdots, s_n]$ as the CNF formula with $m$ clauses where $s_i \in \{-1, 0, 1\}^m$ is an $i$-th propositional variable in $m$-th clauses: 
\begin{gather}
\label{eq:maxsat}
\text{maximize}_{P \in \{-1, 1\}^n} \sum_{i=1}^{m} \bigvee_{j=1}^{n} \bm{1}(s_{ij} p_j > 0).
\end{gather}
We notice that there may not exist any values of propositional variables that satisfy all clauses in the CNF formula to be satisfied. For instance, consider the CNF formula (\ref{eq:cnf}) with propositional variables $p_1$ and $p_2$: there are no assignments of two propositional variables $p_1, p_2$ makes CNF formula satisfiable,
while there exist assignments simultaneously satisfying three clauses out of four clauses exist.

\section{Detailed description of the reasoning layer}
\label{appen:satnet}
As MAXSAT problem in Appendix \ref{appen:maxsat} is a combinatorial optimization problem which is not differentiable, several approaches \citep{wang2018low, goemans1995improved} attempt to relax such MAXSAT problems into the continuous optimization problem. Specifically, they propose the conversion of MAXSAT problem into semidefinite program (SDP). In depth, the  MAXSAT problem in (\ref{eq:maxsat}) can be formulated as the following optimization problem:
\begin{equation}
    \min_{V \in \mathbb{R}^{k \times (n+1)}} <V^TV, S'^TS'> \quad \text{subject to} \;\; ||v_i||=1, \quad i=\top, 1, \cdots, n,
    \label{eq:relax}
\end{equation}
where 
$V:=[v_{\top} \;  v_1 \; \cdots \; v_n] \in \mathbb{R}^{k \times (n+1)}$, 
$S':=[s_{\top} \;  s_1 \; \cdots \; s_m]\text{diag}(1/\sqrt{(4|s_i|})\in \mathbb{R}^{m \times (n+1)}$ with $s_{\top} = \{ -1\}^m$
are relaxations of propositional variables and clauses, respectively. Here, the solution of original MAXSAT problem can be recovered from the solution of SDP in probabilistic manner via randomized rounding, \ie, $\mathbb{P}(p_i = 1) = \cos^{-1}(-v_i^T v_{\top})/\pi$ \citep{goemans1995improved, wang2019satnet}. Moreover, \citet{barvinok1995problems, pataki1998rank} show the optimal solution of the original problem can be recovered from relaxed SDP if $k > \sqrt{2n}$, and \citet{wang2017mixing} 
proves coordinate descent update in this SDP on $V$ converges to the optimal fixed point.

With this continuous relaxation of the MAXSAT problem, \citet{wang2019satnet} proposes SATNet to bridge such continuously relaxed problems and the deep neural network.
Specifically, they regard $S'$ in (\ref{eq:relax}) as the layer weight of the neural network, \ie, they define differentiable forward operation which solves (\ref{eq:relax}) from current weights $S'$ and backward operation to learn a relaxed logical formula from a given dataset via optimizing $S'$. 

\clearpage

\section{Detailed description of the optimization scheme}
\label{appen:optimize}
We first notice that the input $(t_1, \ldots, t_M)$ of the reasoning layer $S_{\psi}$ is non-differentiable, as it includes the $\mathtt{argmax}$ operator. Consequently, optimization of the reasoning loss $\mathcal{L}_{\mathtt{reason}}$ affects the layer parameter $\psi$ but not other parameters, \eg, codebook $\mathcal{E}$. To solve this problem, we propose to use the relaxed version of propositional embeddings $(t_1, \ldots, t_M)$, denoted by $(t_1', \ldots, t_M')^\trans$, where each $(t_i')^\trans = (t_{(i,1)}', \ldots, t_{(i,H)}')^\trans = \in [0, 1]^{H \times K}$ for $\forall i \in [M]$ is defined as follows:
\begin{align*}
    (t_{(i,h)}') \coloneqq \mathtt{softmax} \big(-\lVert z_{(i,h)} - e_1 \rVert_2, \ldots, -\lVert z_{(i,h)} - e_K \rVert_2\big),\quad \forall h \in [H].
\end{align*}
Moreover, we also note that there is no gradient in $\mathcal{L}_{\mathtt{abst}}$ due to the $\mathtt{argmin}$ operator to have $q_i$ from each image $x_i$. To compensate this issue, we simply use straight-through operator \citep{bengio2013estimating}, \ie, we copy gradients of the decoder input $q_i$ to the encoder output $z_i$.

\textbf{Hyperparameters.} We note that \sname contains following hyperparameters: the size of codebook $\mathcal{E}$, the size of spatial features $(H, D)$, the size of reasoning layer $(n,m)$ (see Appendix \ref{appen:satnet}), and the coefficient $\beta$ in the abstraction loss $\mathcal{L}_{\mathtt{abst}}$. In all experiments, we use universal hyperparameter setups: $|\mathcal{E}|=8$, $(H,D)=(25, 192)$, $(n,m)=(1800, 500)$, and $\beta=0.25$.

\section{Details of the RAVEN/i-RAVEN dataset}
\label{appen:raven}
\textbf{RAVEN.}
RAVEN dataset \citep{zhang2019raven} is a synthetic dataset to evaluate the abstract reasoning ability of machines, where each problem is a Raven's Progressive Matrices (RPM) format. Specifically, the dataset consists of total 7 problem types: Center-Single (Center), Left-Right (L-R), Up-Down (U-D), Out-InCenter (O-IC), Out-InGrid (O-ID), 2x2Grid, and 3x3Grid, where each configuration contains 10000 problems. Here, we consider 4 out of total 7 configurations: Center, L-R, U-D, and O-IC. Each image contains five attributes: number of objects, position, shape, size, and color. For rules, the dataset contains 4 rules in total: the attribute is either constant, progressive, arithmetic, and distributed across each row in the problem. Figure \ref{fig:1_strategy} illustrates the Center configuration in RAVEN dataset, and more illustrations are provided in Figure \ref{fig:3_main}.

\textbf{i-RAVEN.}
i-RAVEN dataset \citep{hu2020hierarchical} is a modified version of the RAVEN dataset with a different rule to generate a list of candidates in the problem. To be specific, \citet{hu2020hierarchical} finds there exists a shortcut bias in candidates in RAVEN dataset, \ie, one can find the answer only from candidates without accessing the context images of the problem. Accordingly, they propose a new RAVEN dataset in which such a bias is removed from the candidate set to better measure the reasoning ability of the discriminative abstract reasoning framework. Moreover, they find that the accuracy of existing methods significantly drops if this shortcut bias is removed.
\clearpage

\section{Detailed description of baselines}
\label{appen:baselines}
\textbf{Baselines for qualitative results.} We notice that the neural architecture in \sname is composed of vector-quantized variational autoencoder (VQ-VAE) \citep{van2017neural} and SATNet \citep{wang2019satnet}, based on employing propositional logical prior to solve abstract reasoning problems. To verify the effectiveness of this prior on the abstract reasoning, we qualitatively compare generated answer images from other widely used neural architectures without this assumption. As generative neural architectures for solving reasoning problems are under-explored, we compare the result with other architecture variants where VQ-VAE and SATNet is substituted to different neural architectures.
Specifically, we compare results from combinations of different encoders (autoencoder and VQ-VAE) and reasoning networks (attention layer and 2-layer convolutional neural networks (CNNs) with different kernel sizes, where kernel sizes of CNNs set to 3 and 1).

\textbf{Baselines for qualitative results.}
In the rest of this section, we briefly explain previous approaches to solve the abstract reasoning problem via deep neural networks.
\begin{itemize}[leftmargin=0.2in]
\item \textbf{LSTM} \citep{zhang2019raven} attempts to utilize LSTM to validate the inefficiency of conventional deep neural network architectures to solve reasoning problems.
\item \textbf{WReN} \citep{santoro2018measuring} proposes to solve the abstract reasoning problem via relation network. \citep{NIPS2017_e6acf4b0}.
\item \textbf{LEN} \citep{zheng2019abstract} proposes a variant of relation network, where the input of the network is a triplet of images rather than a pair of images. Furthermore, they empirically verify the performance can be further boosted with the curriculum learning based on a reinforcement learning framework.
\item \textbf{CoPINet} \citep{zhang2019learning} suggests a contrastive learning algorithm to learn underlying rules from given images.
\item \textbf{SRAN} \citep{hu2020hierarchical} designs a hierarchical neural network framework that simultaneously considers images in the problem individually and also at the row and column level.
\end{itemize}
\clearpage

\section{More illustration of generated results}

In this section, we provide additional generated comparisons of different configurations in RAVEN deadset.
\begin{figure}[ht!]
\centering
        \includegraphics[width=0.8\textwidth]{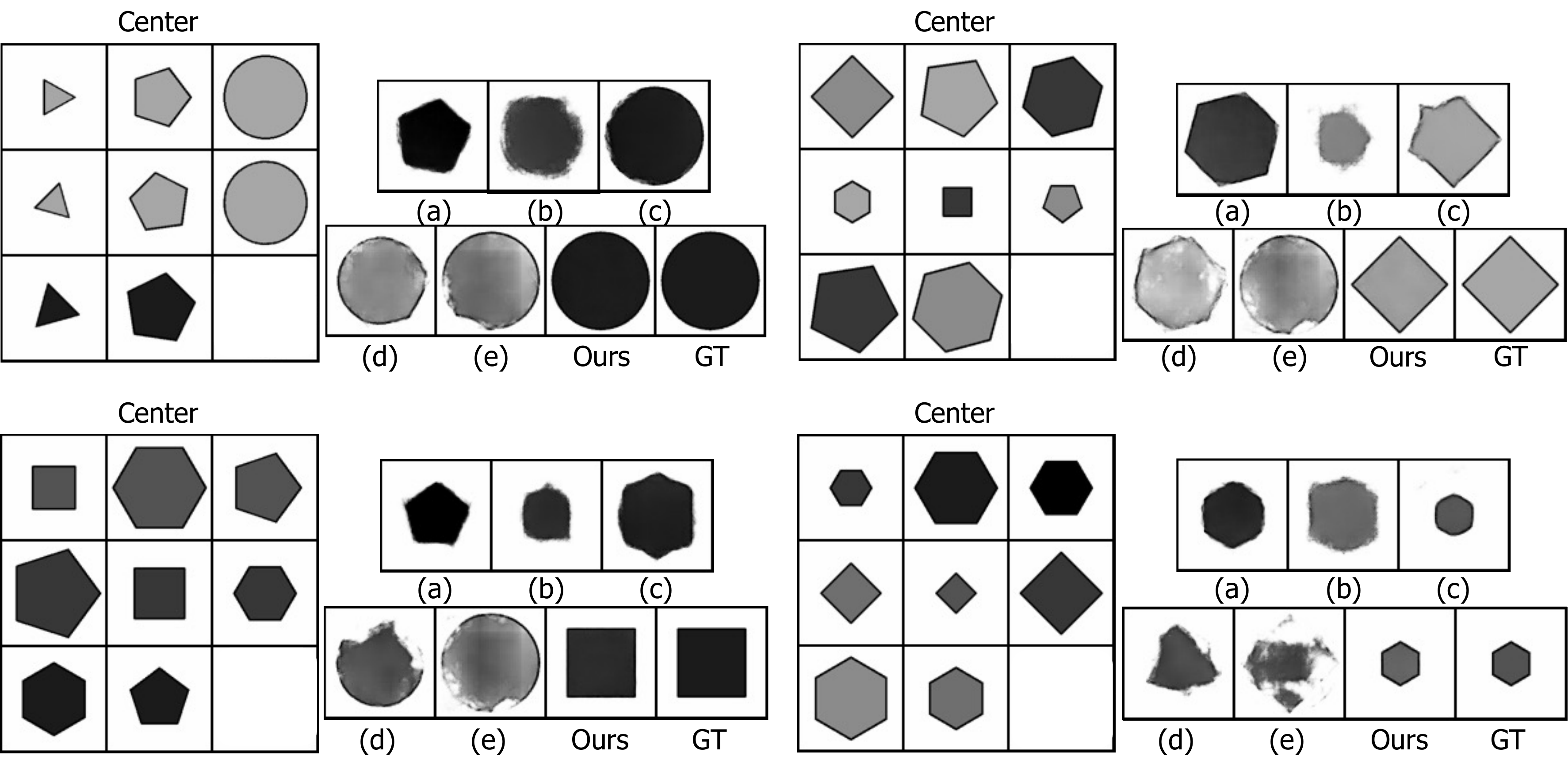}
        \caption{Comparison of generated and ground truth image on Center configuration in RAVEN dataset among different architecture variants. (a)-(c): autoencoder with an attention layer and two convolutional networks with different kernel sizes as the reasoning network. We set the kernel size of 3 and 1 for (b) and (c), respectively. (d)-(e): VQ-VAE with an attention layer and a convolutional network with a kernel size of 3. Ours: our \sname framework, which is composed of VQ-VAE and SATNet. GT: ground-truth answer image.}
        \label{fig:4_cs}
\end{figure}
\begin{figure}[ht!]
\centering
        \includegraphics[width=0.8\textwidth]{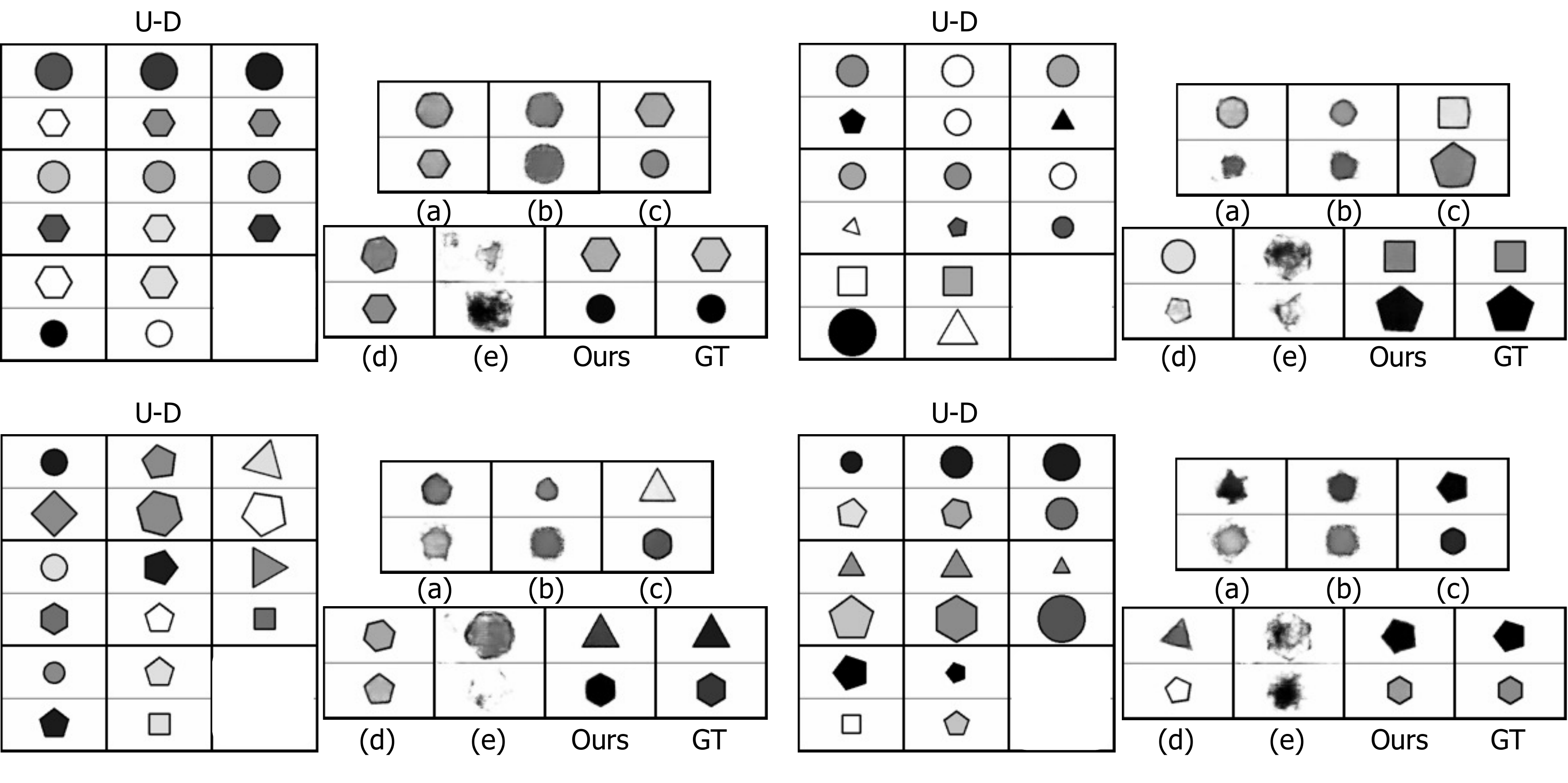}
        \caption{Comparison of generated and ground truth image on U-D configuration in RAVEN dataset among different architecture variants. (a)-(c): autoencoder with an attention layer and two convolutional networks with different kernel sizes as the reasoning network. We set the kernel size of 3 and 1 for (b) and (c), respectively. (d)-(e): VQ-VAE with an attention layer and a convolutional network with a kernel size of 3. Ours: our \sname framework, which is composed of VQ-VAE and SATNet. GT: ground-truth answer image.}
        \label{fig:4_ud}
\end{figure}

\clearpage
\begin{figure}[ht!]
\centering
        \includegraphics[width=0.8\textwidth]{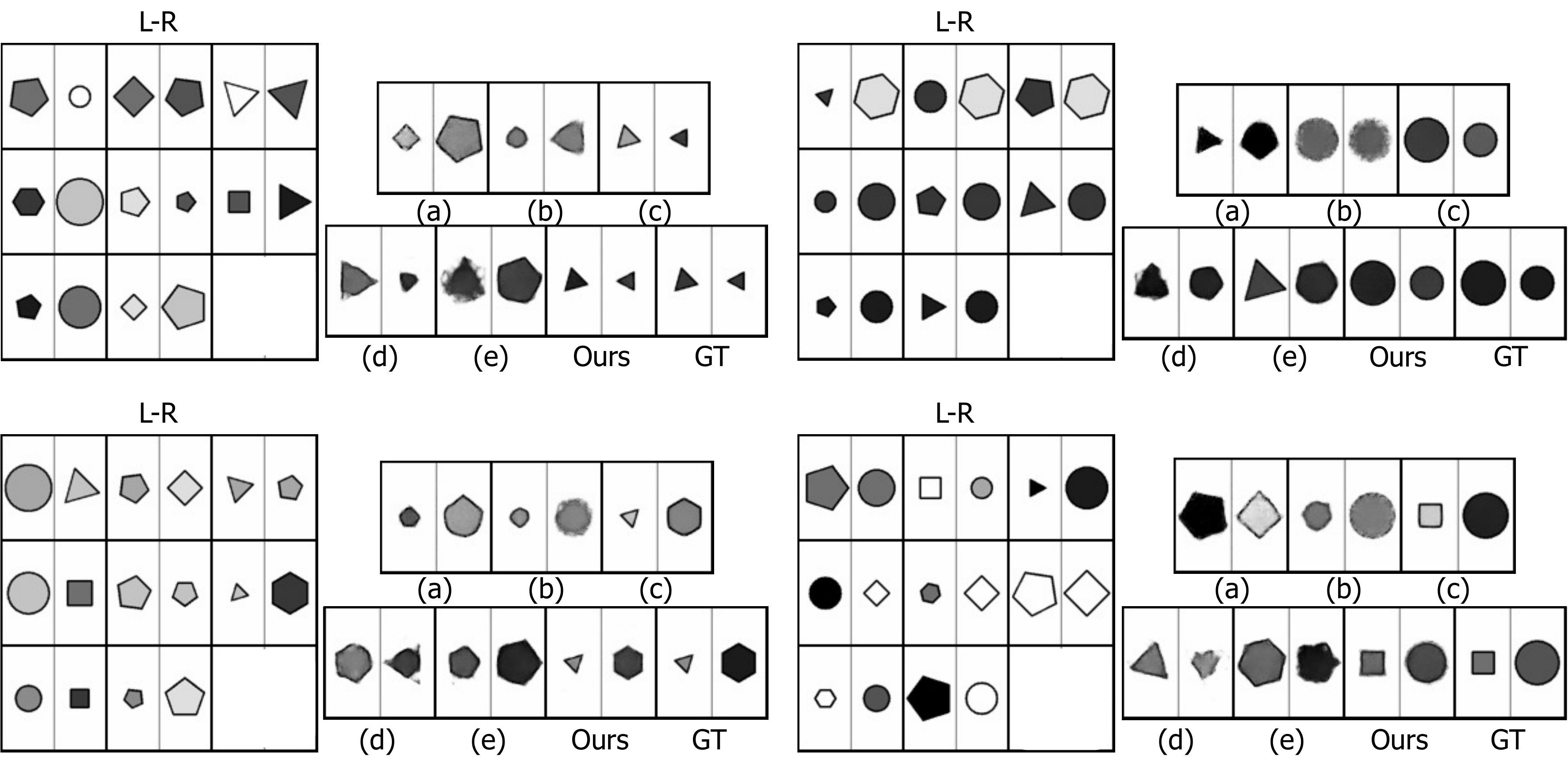}
        \caption{Comparison of generated and ground truth image on L-R configuration in RAVEN dataset among different architecture variants. (a)-(c): autoencoder with an attention layer and two convolutional networks with different kernel sizes as the reasoning network. We set the kernel size of 3 and 1 for (b) and (c), respectively. (d)-(e): VQ-VAE with an attention layer and a convolutional network with a kernel size of 3. Ours: our \sname framework, which is composed of VQ-VAE and SATNet. GT: ground-truth answer image.}
        \label{fig:4_lr}
\end{figure}
\begin{figure}[ht!]
\centering
        \includegraphics[width=0.8\textwidth]{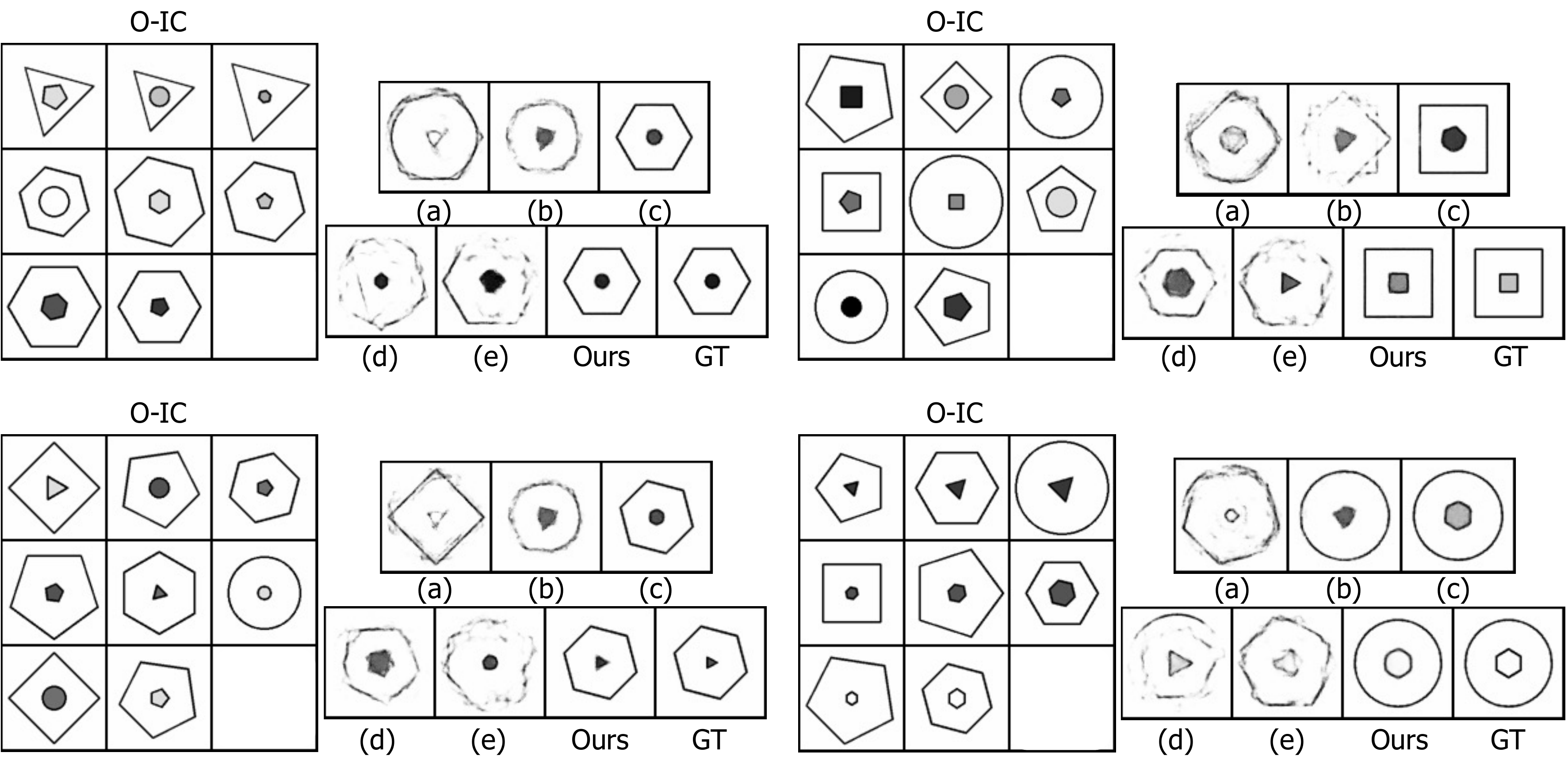}
        \caption{Comparison of generated and ground truth image on O-IC configuration in RAVEN dataset among different architecture variants. (a)-(c): autoencoder with an attention layer and two convolutional networks with different kernel sizes as the reasoning network. We set the kernel size of 3 and 1 for (b) and (c), respectively. (d)-(e): VQ-VAE with an attention layer and a convolutional network with a kernel size of 3. Ours: our \sname framework, which is composed of VQ-VAE and SATNet. GT: ground-truth answer image.}
        \label{fig:4_oic}
\end{figure}

\label{appen:gen}
\clearpage

\end{document}